\documentclass[10pt,twocolumn,letterpaper]{article}

\usepackage{cvpr}
\usepackage{times}
\usepackage{epsfig}
\usepackage{graphicx}
\usepackage{amsmath}
\usepackage{amssymb}
\DeclareMathOperator*{\argmin}{arg\,min}
\DeclareSymbolFont{matha}{OML}{txmi}{m}{it}
\DeclareMathSymbol{\varv}{\mathord}{matha}{118}
\usepackage{array}
\usepackage{tabularx}
\usepackage{multirow, booktabs}
\usepackage{float}

\usepackage[breaklinks=true,bookmarks=false]{hyperref}

\cvprfinalcopy 

\def\cvprPaperID{7726} 

\begin{document}

\title{Fast Symmetric Diffeomorphic Image Registration with Convolutional Neural Networks}

\author{Tony C.W. Mok, Albert C.S. Chung\\
Department of Computer Science and Engineering,\\
The Hong Kong University of Science and Technology\\
{\tt\small cwmokab@connect.ust.hk, achung@cse.ust.hk}
}

\maketitle

	\begin{abstract}
		Diffeomorphic deformable image registration is crucial in many medical image studies, as it offers unique, special properties including topology preservation and invertibility of the transformation. Recent deep learning-based deformable image registration methods achieve fast image registration by leveraging a convolutional neural network (CNN) to learn the spatial transformation from the synthetic ground truth or the similarity metric. However, these approaches often ignore the topology preservation of the transformation and the smoothness of the transformation which is enforced by a global smoothing energy function alone. Moreover, deep learning-based approaches often estimate the displacement field directly, which cannot guarantee the existence of the inverse transformation. In this paper, we present a novel, efficient unsupervised symmetric image registration method which maximizes the similarity between images within the space of diffeomorphic maps and estimates both forward and inverse transformations simultaneously. We evaluate our method on 3D image registration with a large scale brain image dataset. Our method achieves state-of-the-art registration accuracy and running time while maintaining desirable diffeomorphic properties.
	\end{abstract}
	
	
	\section{Introduction}
	Deformable image registration is crucial in a variety of medical imaging studies and has been a topic of active research for decades. The purpose of deformable image registration is to establish the non-linear correspondence between a pair of images and estimate the appropriate non-linear transformation to align a pair of images. This maximizes the customized similarity between the aligned images. Deformable image registration can be useful when analyzing images captured from different sensors, and/or different subjects and different times as it enables the direct comparison of anatomical structures across images from different sources. For example, the manual delineation of anatomical brain structures by an expert is difficult due to the large spatial complexity of an MR brain scan. Also, it usually suffers from the inter-rater variability problem \cite{sparks2002brain}, while deformable image registration enables automatic and robust delineation of brain anatomical structures by registering the target scan to a well-delineated atlas. Traditional deformable registration approaches often model this problem as an optimization problem and strive to minimize the energy function in an iterative fashion. However, this is computationally intensive and time-consuming in practice. Recently, several deep learning-based approaches have been proposed for deformable image registration, which employ a convolutional neural network (CNN) to directly estimate the target displacement field that aligns a pair of input images. Although these methods achieve fast registration and comparable registration accuracy in terms of average Dice score on the anatomical segmentation map, the substantial diffeomorphic properties of the transformation are not guaranteed. In other words, some desirable properties, including topology-preservation and the invertibility of the transformation, for medical imaging studies have been ignored by these approaches.
	
	In this paper, we propose a novel fast symmetric diffeomorphic image registration method that parametrizes the symmetric deformations within the space of diffeomorphic maps using CNN. Specifically, instead of pre-assuming the fixed/moving identity of the input images and outputting a single mapping of all voxels of the moving volume to fixed/target volume, our method learns the symmetric registration function from a collection of $n$-D dataset and output a pair of diffeomorphic maps (with the equivalent length) that map the input images to the middle ground between the images from both geodesic path. Eventually, the forward mapping from one image to another image can be obtained by composing the output diffeomorphic maps and the inverse of the other diffeomorphic map, exploiting the fact that diffeomorphism is a differentiable map and it guarantees there exists a differentiable inverse \cite{avants2008symmetric}.
	
	The main contributions of this work are:
	\begin{itemize}
		\item we present a fast symmetric diffeomorphic image registration method that guarantees topology preservation and invertibility of the transformation;
		\item we propose a novel orientation-consistent regularization to penalize the local regions with negative Jacobian determinant, which further encourages the diffeomorphic property of the transformations; and
		\item our proposed paradigm and objective functions can be transferred to various of applications with minimum effort.
	\end{itemize}
	
	We demonstrate the effectiveness and quality of our method with the example of pairwise registration of 3D brain MR scans. Specifically, we evaluate our method on a large scale T1-weighted MR dataset of over 400 brain scans collected from \cite{marcus2007open}. Results demonstrate that our method not only achieves state-of-the-art registration accuracy, the output transformations are also more consistent with diffeomorphic property as compared with the state-of-the-art deep learning-based registration approaches in both quality and quantitative analysis.
	
	\section{Background}
	\subsection{Deformable registration}
	Image registration refers to the process of warping one (moving) image to align with a second (fixed/reference) image, in which the similarity between the registered images is maximized. Typical transformations, including rigid and affine transformations, allow different degrees of freedom in image transformation and usually serves as an initial transformation for global alignment to deal with large deformation. Deformable image registration is a non-linear registration process that tries to establish the dense voxel-wise non-linear spatial correspondence between fixed/reference image and moving image, which allow much higher degrees of freedom in transformation. Let $F$, $M$ denote the fixed image and the moving image respectively and $\phi$ represents the displacement field. The typical deformable image registration can be formulated as:
	\begin{equation}\label{eq1}
	\phi^{*} = \argmin_{\phi} \mathcal{L}_{sim}(F, M(\phi)) + \mathcal{L}_{reg}(\phi),
	\end{equation}
	
	\noindent where $\phi^*$ denotes the optimal displacement field $\phi$, $\mathcal{L}_{sim}(\cdot,\cdot)$ denotes the dissimilarity function and $\mathcal{L}_{reg}(\cdot)$ represents the smoothness regularization function. In order words, the optimization problem of deformable image registration aims to minimize the dissimilarity (or maximize the similarity) of the fixed image $F$ and warped image $M(\phi)$ while maintaining a smooth deformation field $\phi$. In most of the deformable image registration settings, the affine and scaling transformations have been factored such that the only source of misalignment between the images is non-linear. We follow this assumption throughout this paper. All the brain scans tested in the experiments are affinely registered to the MNI152 space \cite{fonov2011unbiased} in the preprocessing phase. 
	
	\subsection{Diffeomorphic Registration}
	Recent deformable registration approaches often parameterize the deformable model using a displacement field $u$ such that the deformation field $\phi(x) = x + u(x)$, where $x$ denotes the identity transform. Although this parameterization is simple and intuitive, the true inverse transformation of the displacement field is not guaranteed to exist, especially for large and hirsute deformation. Moreover, this deformable model does not necessarily enforce a one-to-one mapping in the transformation. Therefore, throughout this paper, our approach sticks with diffeomorphisms instead. Specifically, we implement our diffeomorphic deformation model with the stationary velocity field. In theory, a diffeomorphism is differentiable and invertible, which guarantees smooth and one-to-one mapping. Therefore, diffeomorphic maps also preserve topology. The path of diffeomorphic deformation fields $\phi_t$ parameterized by $t\in[0, 1]$ can be generated by the velocity fields as:
	\begin{equation}\label{eq2}
	\frac{d\phi_t}{dt} = \textbf{\textit{v}}^{t} (\phi^{t}) = \textbf{\textit{v}}^{t} \circ \phi^{t},
	\end{equation}
	where $\circ$ is a composition operator, $\textbf{\textit{v}}^{t}$ denotes the velocity field at time $t$ and $\phi^{0} = Id$ is the identity transformation. In our settings, the velocity field remains constant over time.
	
	In the literature, the deformation field can be represented as a member of the Lie algebra and is exponentiated to produce a time $1$ deformation $\phi^{(1)}$, which is a member of a Lie group such that $\phi^{(1)} = exp(\textbf{\textit{v}})$. This implies that the exponentiated flow field forces the mapping to be diffeomorphic and invertible using the same flow field. To obtain the time $1$ deformation field $\phi^{(1)}$, we follow \cite{arsigny2006log,ashburner2007fast,dalca2018unsupervised} to integrate the stationary velocity field $\textbf{\textit{v}}$ over time $t=[0, 0.5]$ using the scaling and squaring method for both the fixed image and moving image.  Specifically, given an initial deformation field $\phi^{(1/2^T)} = x + v(x)/2^T$, where $T = 7$ denotes the total time steps we used in our approach. The $\phi^{(1/2)}$ can be obtained using the recurrence $\phi^{(1/2^{t-1})} = \phi^{(1/2^t)} \circ \phi^{(1/2^t)}$, \ie, $\phi^{(1/2)} = \phi^{(1/4)} \circ \phi^{(1/4)}$.
	
	\section{Related Work}
	\subsection{Classic Deformable Registration Methods}
	Classical deformable image registration approaches often optimize a deformation model with constraints iteratively to minimize a custom energy function, which is similar to the optimization problem defined in Eq. \ref{eq1}. Several studies parameterize the problem with displacement fields. The smoothness of the displacement fields is either regularized by an energy function or Gaussian smooth filtering. These methods include Demons \cite{thirion1998image},  free-form deformations with b-splines \cite{rueckert1999nonrigid}, deformable registration via attribute matching and mutual-saliency weighting (DRAMMS) \cite{ou2011dramms}, dense image registration with Markov Random Field \cite{glocker2008dense} and statistical parametric mapping (SPM) \cite{hellier2002inter}. Besides, there are many studies which optimize the registration problem within the space of diffeomorphic maps to ensure the desirable diffeomorphic properties. Popular diffeomorphic registration methods include diffeomorphic Demons \cite{vercauteren2009diffeomorphic}, symmetric image normalization method (SyN) \cite{avants2008symmetric} and diffeomorphic registration using b-splines \cite{rueckert2006diffeomorphic}. These methods often formulate the registration problem as an independent iterative optimization problem. Hence, the registration time increases dramatically, especially when the target image pair contains large variations in anatomical appearance. 
	
	\subsection{Learning-based Deformable Registration Methods}
	Many learning-based approaches, recently, have been proposed for deformable image registration. These approaches often formulate the registration problem as a learning problem with CNN. Recent learning-based methods can be roughly divided into two categories: supervised methods and unsupervised learning methods. 
	Most of the supervised methods \cite{cao2017deformable,rohe2017svf,cao2018deformable,yang2017quicksilver,krebs2017robust} rely on ground truth deformation fields or anatomical segmentation maps to guide the learning process. Although supervised approaches greatly speed up the registration process in the inference phase, the registration accuracy of these methods is bounded by the quality of the synthetic ground truth deformation field or the segmentation map.
	
	Recently, several unsupervised methods have been proposed. These methods utilize a CNN, a spatial transformer and a differentiable similarity function to learn the dense spatial mapping between input images pairs in an unsupervised fashion. Vos et al. \cite{de2017end} demonstrate the efficiency of the unsupervised method with 2D images and adopt cross-correlation as a similarity function. Balakrishnan et al. \cite{balakrishnan2018unsupervised}  generalize the method with 3D volumes and enforce the smoothness of the displacement fields with $L_2$ loss. Dalca et al. \cite{dalca2018unsupervised} proposed a probabilistic diffeomorphic registration method that offers uncertainty estimation. These methods achieve comparable registration accuracy compared to classic registration methods while achieving fast registration.
	
	It is worth noting that most of the existing CNN-based methods parameterize the registration problem with displacement vector fields and ignore the desirable diffeomorphic properties, including topology preservation and the invertibility of the deformation field \cite{cao2017deformable,rohe2017svf,cao2018deformable,yang2017quicksilver,de2017end,balakrishnan2018unsupervised}. Although some methods enforce the smoothness of the displacement field with a global regularization function, it is not sufficient to guarantee that the predicted displacement vectors are smooth and consistent in orientation within the local region. Moreover, the inverse of the transformation is not considered and guaranteed by these methods as well. Specifically, these methods assume the fixed/moving identities of the input images and estimate the transformation from fixed image to moving image.  Motivated by these studies, we present an unsupervised symmetric registration method that is capable of estimating plausible, topology-preserving and inverse-consistent transformations between images from inter-subject.
	
	\section{Method}
	\begin{figure*}
		\begin{center}
            \includegraphics[width=0.8\linewidth]{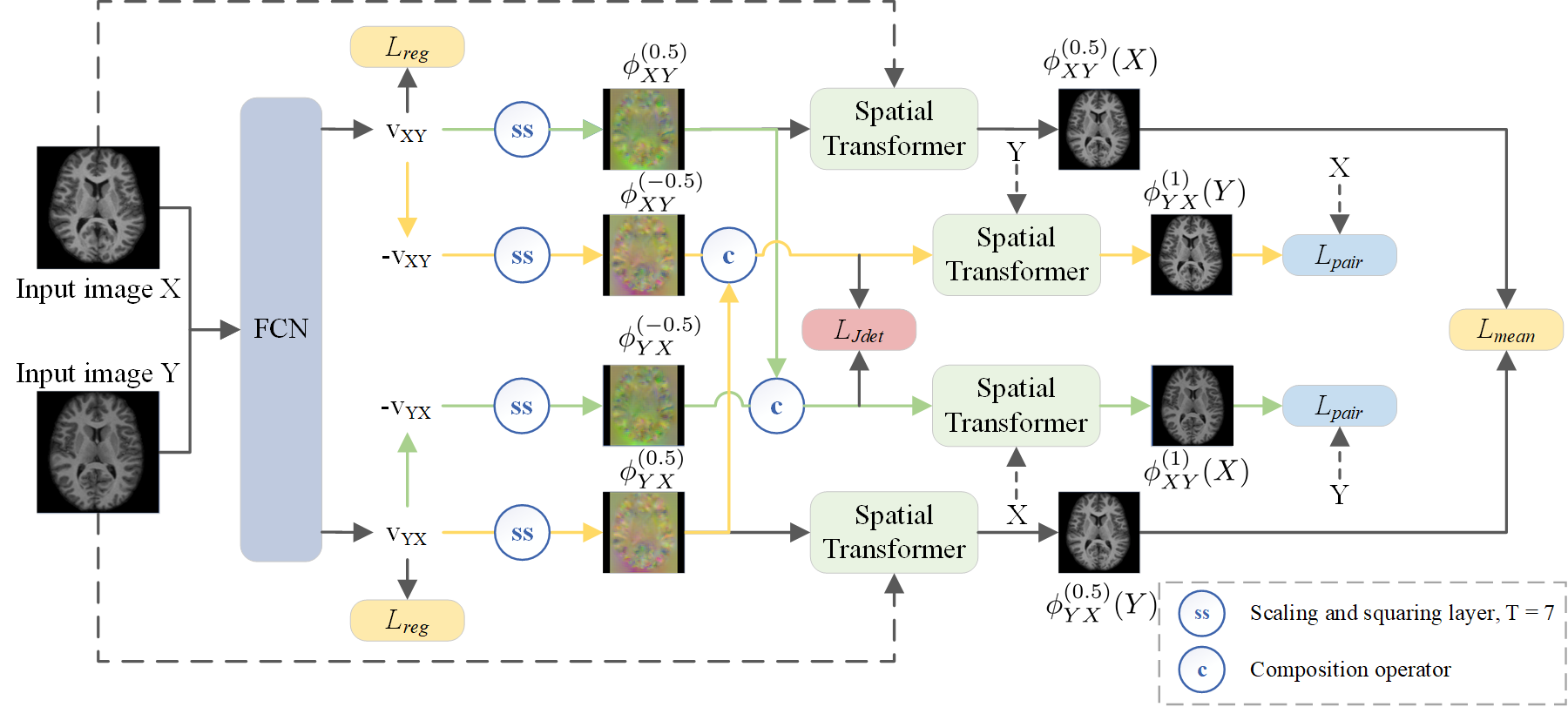}
		\end{center}
		\caption{Overview of the proposed method for symmetric diffeomorphic image registration. We utilize the FCN to learn the symmetric time $0.5$ deformation fields that warp both $X$ and $Y$ to the their mean shape $M$ within the space of diffeomorphic maps. The path with green color depicts the transformation from $X$ to $Y$, while the path with yellow color depicts the transformation from $Y$ to $X$. We omit the magnitude loss $\mathcal{L}_{mag}$ in this figure for simplicity.}
		\label{fig:overview}
	\end{figure*}

    In most of the learning-based deformable image registration approaches, the pair of input images often assigned as a fixed image and a moving image and only one single mapping from the fixed image to the moving image is considered. Moreover, the inverse mapping is often ignored in these approaches. In our symmetric registration settings, we highlight that we do not assume the fixed or moving identity to the input images. Specifically, let $X$, $Y$ be two 3D image volumes defined in a mutual spatial domain $\Omega \subset \mathcal{R}^3$. The deformable registration problem can be parametrized as a function $f_\theta (X, Y) = ( \phi_{XY}^{(1)}, \phi_{YX}^{(1)})$, where $\theta$ denotes the learning parameters in CNN. $\phi_{XY}^{(1)} = \phi_{XY}(x, 1)$ and $\phi_{YX}^{(1)} = \phi_{YX}(y, 1)$ represent the time $1$ diffeomorphic deformation fields that warp the identity position of some anatomical position $x {\in} X$ toward $y {\in} Y$ and warps $y {\in} Y$  toward $x {\in} X$ respectively. Motivated by the conventional non-learning based symmetric image normalization methods \cite{wu2012hierarchical,avants2008symmetric,reaungamornrat2016mind}, we propose to learn the two separated time $0.5$ deformation fields that warp both $X$ and $Y$ to their mean shape $M$ in the geodesic path. After the model converges,  the time $1$ deformation fields that warp $X$ to $Y$ and $Y$ to $X$ can be obtained by the composition of two estimated time $0.5$ deformation fields subject to the fact that diffeomorphism is a differentiable map and it guarantees a differentiable inverse exists \cite{ashburner2007fast}. The transformation from $X$ to $Y$ is decomposed into $\phi_{XY}^{(1)} = \phi_{YX}^{(-0.5)}(\phi_{XY}^{(0.5)}(x))$, while the transformation from $Y$ to $X$ is decomposed into $\phi_{YX}^{(1)} = \phi_{XY}^{(-0.5)}(\phi_{YX}^{(0.5)}(y))$. Hence, the function $f_\theta$ can be rewritten as $f_\theta (X, Y) = ( \phi_{YX}^{(-0.5)}(\phi_{XY}^{(0.5)}(x)), \phi_{XY}^{(-0.5)}(\phi_{YX}^{(0.5)}(y)))$.
	
	\subsection{Symmetric Diffeomorphic Neural Network}
	As shown in Fig. \ref{fig:overview}, we parametrized the function $f_\theta$ using a fully convolutional neural network (FCN), several scaling and squaring layers and differentiable spatial transformers \cite{jaderberg2015spatial}. $\phi_{XY}^{(0.5)}$ and $\phi_{YX}^{(0.5)}$ are computed using the scaling and squaring method with the estimated velocity fields $\textbf{\textit{v}}_{XY}$ and $\textbf{\textit{v}}_{YX}$ respectively.
	
	\begin{figure}
		\begin{center}
            \includegraphics[width=1.0\linewidth]{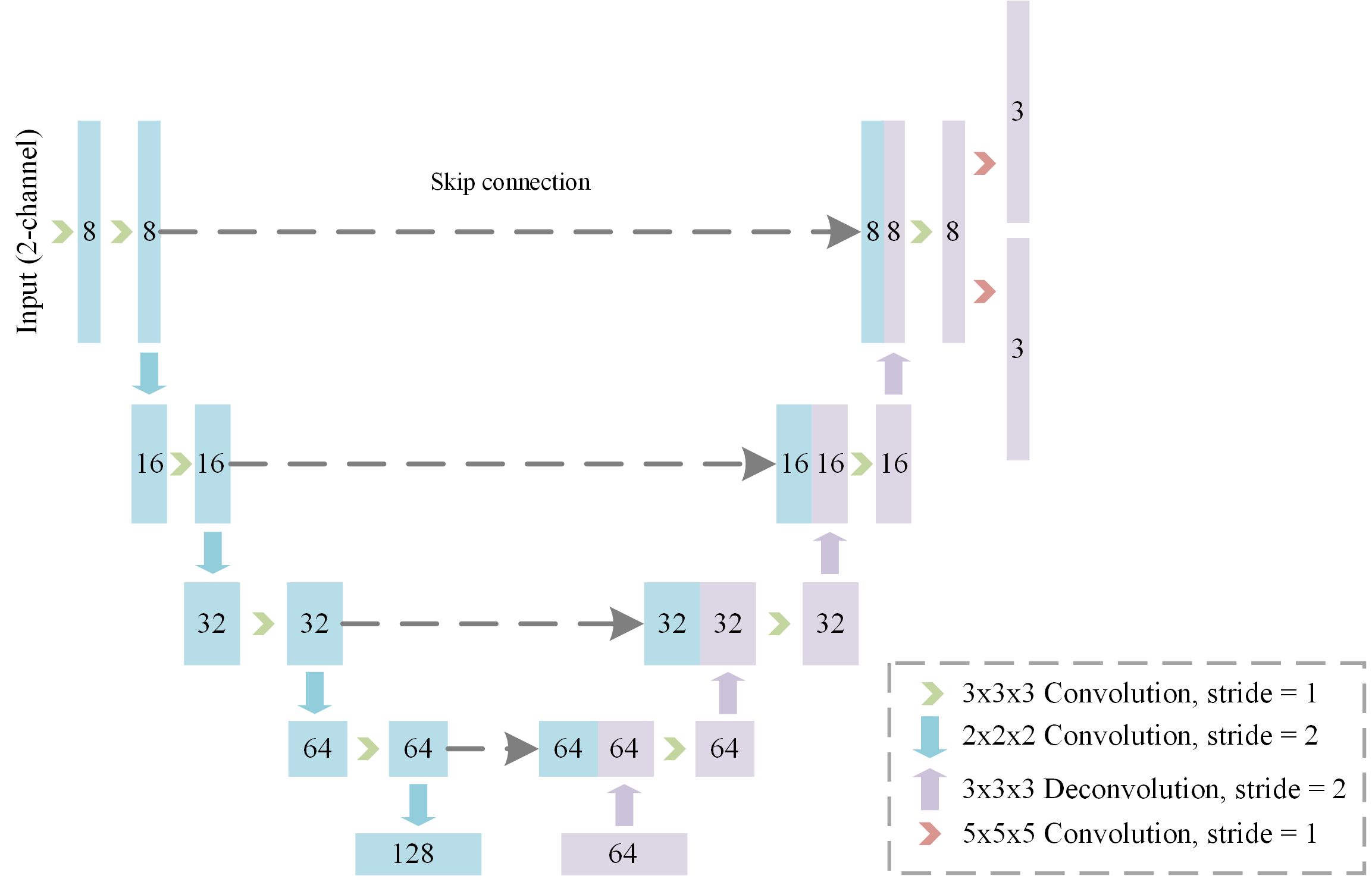}
		\end{center}
		\caption{An illustration of the proposed fully convolutional networks architecture that utilized to estimate the target velocity fields $\textbf{\textit{v}}_{XY}$ and $\textbf{\textit{v}}_{YX}$. The blocks highlighted with blue and purple color indicate the 3D feature maps from the encoder and decoder respectively.}
		\label{fig:FCN_archi}
 		\vspace{-14pt}
	\end{figure}
		
		
    The architecture of our FCN is similar to U-Net \cite{ronneberger2015u}, which consists of an 5-level hierarchical encoder-decoder with skip connections as shown in Fig. \ref{fig:FCN_archi}. The proposed FCN concatenates $X$ and $Y$ as a single 2-channels input and learns to estimate two dense, non-linear velocity fields $\textbf{\textit{v}}_{XY}$ and $\textbf{\textit{v}}_{YX}$ from $X$ and $Y$ jointly from the beginning. For each level in the encoder, we apply two successive convolution layers, which contain one $3 \times 3 \times 3$ convolution layer with a stride of $1$, followed by a $3 \times 3 \times 3$ convolution layer with a stride of $2$ to further compute the high-level features between the inputs and to downsample the features in half until the lowest level is reached. For each level in the decoder, we concatenate the feature maps from the encoder through skip connection and apply $3 \times 3 \times 3$ convolution with a stride of $1$ and $2 \times 2 \times 2$ deconvolution layer for upsampling the feature maps to twice of its size. At the end of the decoder, two $5 \times 5 \times 5$ convolution layers with a stride of $1$ are appended to the last convolution layer and generate the velocity fields $\textbf{\textit{v}}_{XY}$ and $\textbf{\textit{v}}_{YX}$, followed by a \emph{softsign} activation function (\ie, $SoftSign(x)=\frac{x}{1+|x|}$). It then multiplies itself by a constant $c$, to normalize the velocity fields within the range $[-c, c]$. We set $c = 100$ such that it is sufficient for large deformation. Empirically, the non-linear misalignment is usually less than $25$ voxels in the deformable registration of brain MR scans with $1mm^3$ resolution. In our FCN, each convolution layer is followed by a rectified linear unit (ReLU) activation, except for the output convolution layers.
    
    Besides, we follow \cite{arsigny2006log,dalca2018unsupervised} to implement the scaling and squaring layer with a differentiable spatial transformer and utilize it to integrate the estimated velocity fields to time $0.5$ deformation fields $\phi_{XY}^{(0.5)}$ and $\phi_{YX}^{(0.5)}$, subject to $\phi^{(1)} = exp(\textbf{\textit{v}})$. Specifically, given a constant time step $T$, we initialize $\phi_{XY}^{(1/2^T)} = x + \textbf{\textit{v}}_{XY}(x)/2^T$ and $\phi_{YX}^{(1/2^T)} = x + \textbf{\textit{v}}_{YX}(x)/2^T$. We compute the time 0.5 deformation fields through the recurrence $\phi^{(1/2^{t-1})} = \phi^{(1/2^t)} \circ \phi^{(1/2^t)}$  until $t=2$. The composition of two deformation fields is computed using a differentiable spatial transformer with trilinear interpolation such that $\phi^{(1/t)} \circ \phi^{(1/t)} = \phi^{(1/t)}(\phi^{(1/t)} (x))$. Since the deformation fields are diffeomorphic and the mapping is one-to-one, we exploit the fact that the inverse transformations can be computed by integrating the same velocity field backward, such that $\phi^{(-1/2^T)} = x - \textbf{\textit{v}}(x)/2^T$ and the recurrence denoted as $\phi^{(-1/2^{t-1})} = \phi^{(-1/2^t)} \circ \phi^{(-1/2^t)}$.
    
    Moreover, a spatial transformer is utilized to transform the image based on the input image and the computed deformation field. Specifically, we implement the spatial transformer with an identity grid generator and trilinear sampler.  The deformation field computed by the scaling and squaring layer is added to the identity grid. Then, the trilinear sampler uses the resulting grid to warp the input image. In particular, the spatial transformer generates the warped images $X(\phi_{XY}^{(0.5)})$, $Y(\phi_{YX}^{(0.5)})$, $X(\phi_{XY}^{(1)})$ and $Y(\phi_{YX}^{(1)})$ with the estimated deformation field $\phi_{XY}^{(0.5)}$, $\phi_{YX}^{(0.5)}$, $\phi_{YX}^{(-0.5)}(\phi_{XY}^{(0.5)}(x)$ and $\phi_{XY}^{(-0.5)}(\phi_{YX}^{(0.5)}(y))$ respectively, as shown in Fig. \ref{fig:overview}.
		
	
	\subsection{Symmetric Similarity}
	Existing CNN-based methods often ignore desirable diffeomorphic properties, including topology preservation, invertibility and inverse consistency of the transformation \cite{cao2017deformable,rohe2017svf,cao2018deformable,yang2017quicksilver,de2017end,balakrishnan2018unsupervised}. Inspired by the classic iterative-based symmetric normalization methods \cite{wu2012hierarchical,avants2008symmetric,reaungamornrat2016mind}, our method estimates the transformations (\eg, $\phi_{XY}^{(0.5)}$ and $\phi_{YX}^{(0.5)}$) from both $X$ and $Y$ to the mean shape $M$, and the transformations (\eg, $\phi_{XY}^{(1)}$ and $\phi_{YX}^{(1)}$) that warp $X$ to $Y$ and $Y$ to $X$. We propose to minimize the symmetric mean shape similarity loss $\mathcal{L}_{mean}$ and pairwise-similarity loss $\mathcal{L}_{sim}$ by gradient descent, which enforce the invertibility and the inverse consistency of the predicted transformations. Similar to the existing CNN-based methods, our proposed method is compatible with any differentiable similarity metrics such as normalized cross-correlation (NCC), mean squared error (MSE), sum of squares distance (SSD) and mutual information (MI). For simplicity, we utilize the normalized cross-correlation NCC as our similarity metric to compute the degree of alignment between two images. Let $I$ and $J$ be two input image volumes, $\bar{I}(x)$ and $\bar{J}(x)$ be the local mean of $I$ and $J$ at position $x$ respectively. The local mean is computed over a local $w^3$ window centered at each position $x$, with $w=7$ in our experiments. The NCC is defined as follows:
	\begin{equation}\label{eq:ncc}
	\begin{split}
	& NCC(I, J) = \\
	& \sum_{x \in \Omega} \frac{\sum_{x_i} (I(x_i) - \bar{I}(x)) (J(x_i) - \bar{J}(x))}{\sqrt{\sum_{x_i} (I(x_i) - \bar{I}(x))^2 \sum_{x_i} (J(x_i) - \bar{J}(x))^2}},
	\end{split}
	\end{equation}
	where $x_i$ denotes the position within $w^3$ local windows centered at $x$.
	
	Specifically, our proposed similarity loss function $\mathcal{L}_{sim}$ consists of two symmetric loss terms: mean shape similarity loss $\mathcal{L}_{mean}$ and pairwise similarity loss $\mathcal{L}_{pair}$.  The $\mathcal{L}_{mean}$ measures the dissimilarity between the warped $X$ and warped $Y$, which toward the mean shape $M$, while the $\mathcal{L}_{pair}$ measures the pairwise dissimilarity between the warped $X$ to $Y$ and warped $Y$ to $X$. The proposed similarity loss function is then formulated as: 
	
	\begin{equation}\label{eq:sim_loss}
	\mathcal{L}_{sim} = \mathcal{L}_{mean} + \mathcal{L}_{pair}
	\end{equation}
	
	\noindent with
	
	\begin{equation}\label{eq:sim_loss_mean}
	\mathcal{L}_{mean} = -NCC(X(\phi_{XY}^{(0.5)}), Y(\phi_{YX}^{(0.5)}))
	\end{equation}
	
	\noindent and
	
	\begin{equation}\label{eq:sim_loss_pair}
	\mathcal{L}_{pair} = -NCC(X(\phi_{XY}^{(1)}), Y) - NCC(Y(\phi_{YX}^{(1)}), X)
	\end{equation}
	where $\phi_{XY}^{(1)}$ (and $\phi_{YX}^{(1)}$) can be decomposed into $\phi_{YX}^{(-0.5)} \circ \phi_{XY}^{(0.5)}$ (and $\phi_{XY}^{(-0.5)} \circ \phi_{YX}^{(0.5)}$) in diffeomorphic space.
	In other words, minimizing the $\mathcal{L}_{sim}$ tends to maximize the similarity of the warped images in a bidirectional fashion. Furthermore, not only does our method inherit the topology-preservation and invertibility properties from the diffeomorphic deformation model, the inverse consistency is implicitly guaranteed by the proposed pairwise similarity loss function as it considers the transformation from both directions.
	
	\subsection{Local Orientation Consistency}
	Existing learning-based approaches \cite{balakrishnan2018unsupervised,de2019deep,kim2019unsupervised} often regularize the deformation field with a regularization loss function, such as an $L_2$-norm on the spatial gradients of the deformation field. Although the smoothness of the deformation field can be controlled by the weight of the regularizer, the global regularizer may greatly degrade the registration accuracy of the model, especially when a large weight is assigned for the regularizer. Furthermore, these regularizers are not sufficient to secure a topology-preservation transformation in practice. To address this issue, we propose a novel selective Jacobian determinant regularization that imposes a local orientation consistency constraint on the estimated deformation field. Mathematically, the proposed selective Jacobian determinant regularization loss $\mathcal{L}_{Jdet}$ is defined as: 
    \begin{equation}\label{eq:loss_jdet}
	\mathcal{L}_{Jdet} = \frac{1}{N} \sum_{p \in \Omega} \sigma (-|J_{\phi}(p)|),
	\end{equation}
	where $N$ denotes the total number of elements in $|J_\phi|$, $\sigma(\cdot)$ represents an activation function that is linear for all positive values and zero for all negative values. In our experiments, we set $\sigma(\cdot) = max(0,\cdot)$, which is equivalent to the ReLU function and $|J_\phi(\cdot)|$ denotes the determinant of the Jacobian matrix deformation field $\phi$ at position $p$. The definition of Jacobian matrix $J_\phi(p)$ can be written as: 
	
	\begin{equation}\label{eq:jdet}
	J_{\phi}(p) = 
	\begin{pmatrix}
	\frac{\partial \phi_x(p)}{\partial x} & \frac{\partial \phi_x(p)}{\partial y} & \frac{\partial \phi_x(p)}{\partial z} \\ 
	\frac{\partial \phi_y(p)}{\partial x} & \frac{\partial \phi_y(p)}{\partial y} & \frac{\partial \phi_y(p)}{\partial z} \\ 
	\frac{\partial \phi_z(p)}{\partial x} & \frac{\partial \phi_z(p)}{\partial y} & \frac{\partial \phi_z(p)}{\partial z}
	\end{pmatrix}
	\end{equation}
	The Jacobian matrix of the deformation fields is a second-order tensor field formed by the derivatives of the deformations in each direction. The determinant of the Jacobian determinant could be useful in analyzing the local behavior of the deformation field. For example, a positive point $p \in |J_\phi|$ means the deformation field at point $p$ preserves orientation in the neighborhood of $p$. On the contrary, if the point $p \in |J_\phi|$ is negative, the deformation field at point $p$ reverses the orientation in the neighborhood of $p$ and, hence, the one-to-one mapping has been lost. We exploit this fact to enforce the local orientation consistency on the deformation fields by penalizing the local region with a negative Jacobian determinant, while the region with positive Jacobian determinant (\ie, consistence orientation in the neighborhood) will not be affected by this regularization loss. It is worth noting that the proposed selective Jacobian determinant regularization loss means not to replace the global regularizer. Instead, we utilize both regularization loss functions in our method to produce smooth and topology-preservation transformations while alleviating the tradeoff between smoothness and registration accuracy. In particular, we further enforce the smoothness of the velocity fields with $\mathcal{L}_{reg} = \sum_{p \in \Omega}(||\nabla \textbf{\textit{v}}_{XY}(p)||^2_2 + ||\nabla \textbf{\textit{v}}_{YX}(p)||^2_2)$.
	
    Besides, we further avoid the bias on either path by imposing a magnitude constraint $\mathcal{L}_{mag} = \frac{1}{N}(||\textbf{\textit{v}}_{XY}||^2_2 - ||\textbf{\textit{v}}_{YX}||^2_2)$, which  explicitly guarantees the magnitude of the predicted velocity fields are (approximately) the same.

    Therefore, the complete loss function of our method can be written as:
	\begin{equation}
	\mathcal{L}(X, Y) = \mathcal{L}_{sim} + \lambda_{1} \mathcal{L}_{Jdet} + \lambda_{2} \mathcal{L}_{reg} + \lambda_{3} \mathcal{L}_{mag},
	\end{equation}
	where $\lambda_{1}$, $\lambda_{2}$ and $\lambda_{3}$ are the weights to balance the contributions of the orientation consistency loss, regularization loss, and magnitude loss respectively.
	
	\section{Experiments}
	
	\subsection{Data and Pre-processing}
	We evaluated our method on brain atlas-based registration using 425 T1-weighted brain MRI scans from OASIS \cite{marcus2007open} dataset. Subjects aged from 18 to 96 and 100 of the included subjects have been clinically diagnosed with very mild to moderate Alzheimer’s disease. We resampled all MRI scans to $256 \times 256 \times 256$ with the same resolution ($1mm \times 1mm \times 1mm$) followed by standard preprocessing steps, including motion correction, skull stripping, affine spatial normalization and subcortical structures segmentation, for each MRI scan using FreeSurfer \cite{fischl2012freesurfer}. Then, we center cropped the resulting MRI scan to $144 \times 192 \times 160$. Subcortical segmentation maps, including 26 anatomical structures, serve as the ground truth to evaluate our method. We split the dataset into 255, 20 and 150 volumes for train, validation and test sets respectively.
    We evaluate our method on the atlas-based registration task. Atlas-based registration is a common application in analyzing inter-subject images, which aims to establish the anatomical correspondence between the atlas and the target image (moving image). The atlas could be a single volume or the average image volume among images within the same space. In our experiments, we randomly select 5 MR volumes from the test set as the atlas and we perform atlas-based registration with different deformable registration approaches, which align the reminding image volumes in the test set to match the selected atlas. Hence, we register 725 pairs of volumes in the test set for each method in total. During the evaluation, we set $X$ to atlas and $Y$ to the moving subject for our method.
	\begin{figure}[H]
		\begin{center}
		\includegraphics[width=1.0\linewidth]{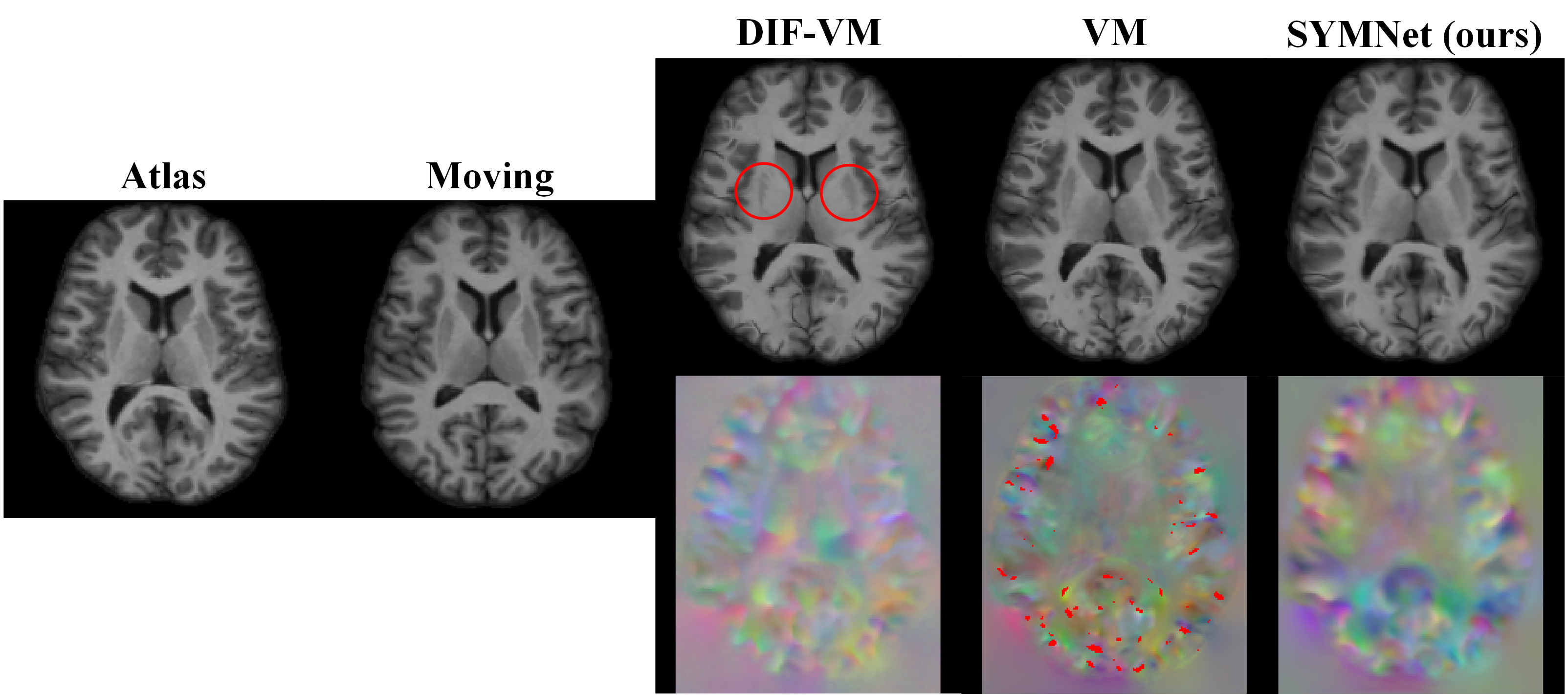}
		\end{center}
		\caption{Example axial MR slices from the atlas, moving image, resulting warped image and deformation field for DIF-VM, VM and our method. The region with non-positive Jacobian determinant in each deformation field is overlayed with red color. The circles in red color highlight the artifact on the left and right putamen from the result of DIF-VM.}
		\label{fig:qa}
	\end{figure}
	
	\subsection{Measurement}
	Since the ideal ground truth of the non-linear deformation field is not well-defined, we evaluate a registration algorithm with two common metrics, Dice similarity coefficient (DSC) and Jacobian determinant ($|J_\phi|$). Specifically, we first register each brain MR volume to an atlas. Then, we warp the anatomical segmentation map of the subject to align with the atlas segmentation map using the resulting deformation fields. Subsequently, we evaluate the overlap of the segmentation maps using DSC and the diffeomorphic property of the predicted deformation fields using the Jacobian determinant.
	
	\subsubsection{Dice Similarity Coefficient (DSC)}
\begin{figure*}[t!]
		\centering
		\includegraphics[width=0.7\linewidth]{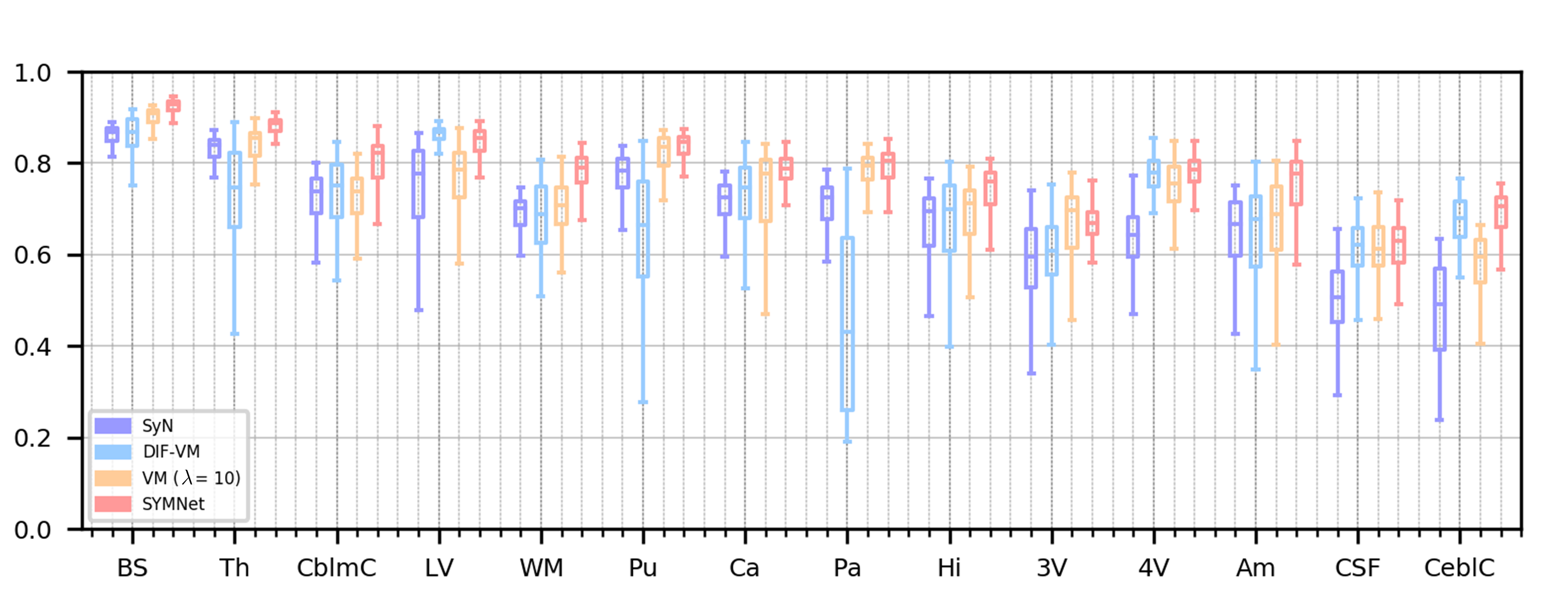}
		\vspace{-2pt}
		\caption{Boxplots illustrating Dice scores of each anatomical structure for SyN, DIF-VM, VM($\lambda=10$) and our method. Left and right brain hemispheres are combined into one structure for visualization. Brain stem (BS), thalamus (Th), cerebellum cortex (CblmC), lateral ventricle (LV), cerebellum white matter (WM), putamen (Pu), caudate (Ca), pallidum (Pa), hippocampus (Hi), 3rd ventricle (3V), 4th ventricle (4V), amygdala (Am), CSF (CSF), and cerebral cortex (CeblC) are included.}
		\label{fig:boxplot}
	\end{figure*}

    DSC measures the spatial overlap of anatomical segmentation maps between the atlas and warped moving volume. In particular, 26 anatomical structures were included in our analysis as shown in Fig. \ref{fig:boxplot}. The value of DSC ranges from $[0,1]$ and a well-registered moving MRI volume should show a high anatomical correspondence to the atlas, and hence yielding a high DSC score.
	
	\subsubsection{Jacobian Determinant}
	Jacobian matrix is the derivatives of the deformations, which captures the local behaviors of the deformation field, including shearing, stretching and rotating of the deformation field. The definition of the Jacobian matrix $J_\phi(p)$ is defined in eq \ref{eq:jdet}. In theory, the local deformation field is diffeomorphic, including topology-preserving and invertible, only for the regions with positive Jacobian determinant (\ie, $|J_\phi(p)|>0$). In contrast, local regions with negative Jacobian determinant indicate that the one-to-one mapping has been lost. In our experiments, we compute the Jacobian determinant of the deformation fields and count the number of voxels with non-positive Jacobian determinant (\ie, $|J_\phi(p)| \leq 0$).
	
	\subsection{Baseline Methods}
    We compare our proposed method to the classic symmetric image normalization method (SyN) \cite{avants2008symmetric} and two unsupervised learning-based deformable registration methods \cite{balakrishnan2018unsupervised,dalca2018unsupervised}, denoted as VM and DIF-VM. SyN is one of the top-performing registration algorithms among 14 typical nonlinear deformation algorithms \cite{klein2009evaluation}. VM and DIF-VM are the cutting edge unsupervised deformable registration methods proposed recently. VM utilizes a CNN and a diffusion regularizer to estimate displacement vector fields while DIF-VM presents a probabilistic diffeomorphic registration method with CNN. For SyN, we use the SyN implementation in the ANTs package \cite{avants2011reproducible} with careful parameter tuning. Since SyN is an iterative-based approach, we set the maximum iteration to $(200, 100, 50)$ for each level to balance the tradeoff between registration accuracy and running time. For the learning-based methods (VM and DIF-VM), we used their official implementation online (https://github.com/voxelmorph/voxelmorph), which is developed and maintained by the authors. We train VM and DIF-VM from scratch and followed the optimal parameters setting in \cite{balakrishnan2018unsupervised,dalca2018unsupervised} to obtain the best performance. Different from the experiment settings in \cite{balakrishnan2018unsupervised,dalca2018unsupervised}, we train learning-based methods by pairwise registration with image volume pairs in training set only, and hence, the atlases are not included in the training phase. Also, to study the effect of the regularizer, we train VM with different weights for the regularizer.
	
	\subsection{Implementation}
	Our proposed method (denoted as SYMNet) is implemented based on Pytorch \cite{paszke2017automatic}. We adopt the stochastic gradient descent (SGD) \cite{bottou2010large} optimizer with the learning rate and momentum set to $1e^{-4}$ and $0.9$ respectively. We obtain the best result with $\lambda_1=1000$, $\lambda_2=3$ and $\lambda_3=0.1$. All the parameters were tuned by grid search. We train our network on a GTX 1080Ti GPU and select the model that obtaining the highest Dice score on the validation set. To evaluate the effectiveness of the proposed local orientation consistency loss, we compare SYMNet to its variant (denotes as SYMNet-1), in which the proposed local orientation consistency loss is removed during the training phase.
	
	\begin{table}
		\centering	
		\renewcommand{\arraystretch}{0.95}
		\newcolumntype{C}[1]{>{\centering\arraybackslash}p{#1}}
		\begin{tabular}{C{2.2cm} C{2cm} C{2.8cm}}
			\toprule
			Method &  Avg. DSC & $|J_\phi|\leq0$\\
			\midrule
			Affine &  0.567 (0.180) & - \\
			\midrule
			SyN &  0.680 (0.132) & 0.047 (0.612) \\
			\midrule
			DIF-VM & 0.693 (0.156) & 346.712 (703.418)\\
			\midrule
			VM ($\lambda=1$) & 0.727 (0.144) & 116168 (88739)\\
			VM ($\lambda=5$) & 0.712 (0.132) & 266.594 (246.811)\\
			VM ($\lambda=10$) & 0.707 (0.128)  & 0.588 (0.764)\\
			\midrule
			SYMNet-1 & 0.743 (0.113) & 1156 (2015) \\
			SYMNet &  0.738 (0.108) & 0.471 (0.921) \\
			\bottomrule
		\end{tabular}
		\vspace{8pt}
		\caption{Average Dice scores (higher is better) and average number of voxels with non-positive Jacobian Determinant (lower is better). Standard deviations are shown in parentheses. Affine: Affine spatial normalization.}\label{tab:result}
		\vspace{-8pt}
	\end{table}
	
	\subsection{Results}
	\subsubsection{Registration Performance}
	Table \ref{tab:os_loss} shows average DSC and number of voxels with non-positive Jacobian determinant over all subjects and structures for a baseline of affine normalization, SyN, DIF-VM, VM (and its variants), and our proposed method SYMNet. All the learning-based methods (DIF-VM, VM and SYMNet) outperform SyN in terms of average DSC. However, VM does not yield diffeomorphic results since the number voxels with non-positive Jacobian determinant is significantly large. Fig. \ref{fig:qa} shows an example axial MR slices from resulting warped image for DIF-VM, VM and our method. Although DIF-VM reports comparable registration accuracy with VM in \cite{dalca2018unsupervised}, we found that resulting warped image from DIF-VM is often sub-optimal, especially in left and right Putamen. Also, we observe that the resulting deformation fields from VM are discontinuous. We visualize the regions with non-positive Jacobian determinant with red color in the resulting deformation fields. Our proposed method achieves the overall best performance in terms of average DSC, while maintaining the number voxels with non-positive Jacobian determinant close to zero, which implies that our resulting deformation fields guarantee the desirable diffeomorphic properties. The boxplots in Fig. \ref{fig:boxplot} illustrate the distribution of DSC for each anatomical structure. Compare to methods with diffeomorphic properties, our proposed method achieves the best performance in all anatomical structures over all the methods.
	
	
	\subsubsection{Effect of the Local Orientation-consistent Loss}
    \begin{table}
		\centering	
		\renewcommand{\arraystretch}{0.95}
		\newcolumntype{C}[1]{>{\centering\arraybackslash}p{#1}}
		\newcolumntype{L}[1]{>{\arraybackslash}p{#1}}
		\newcommand*{\MyIndent}{\hspace*{0.35cm}}%
		\begin{tabular}{L{2.2cm} C{2.6cm} C{2.2cm}}
			\toprule
			\MyIndent$\lambda_1$ &  Avg. DSC & $|J_\phi|\leq0$\\
			\midrule
			\MyIndent$\lambda_1=0$ &  0.7434 (0.113) & 1156 (2015)\\
			\MyIndent$\lambda_1=1$ &  0.7431 (0.110) & 860 (1562) \\
			\MyIndent$\lambda_1=10$ &  0.7423 (0.111) & 460 (845)\\
			\MyIndent$\lambda_1=100$ & 0.7408 (0.104) & 133 (260)\\
            \MyIndent$\lambda_1=1000$ & 0.7381 (0.108) & 0.471 (0.921)\\
			\bottomrule
		\end{tabular}
		\vspace{8pt}
		\caption{Influence of the proposed local orientation consistency loss with varying weights. Average Dice scores (higher is better) and average number of voxels with non-positive Jacobian Determinant (lower is better). Standard deviations are shown in parentheses.}\label{tab:os_loss}
		\vspace{-8pt}
	\end{table}

    Table \ref{tab:os_loss} presents the effect of the proposed local orientation-consistent loss on DSC and the number of voxels with $|J_{\phi}| <= 0$ with varying weights $\lambda_1$. Although both our method (SYMNet-1) and DIF-VM optimize the problem in diffeomorphic space, the experiments in Table \ref{tab:result} show that the resulting solutions are not necessary diffeomorphic. The underlying reasons are that the deformations can only be represented discretely with a finite number of parameters and the interpolations used during the integration of velocity fields could cause violations. The results in Table \ref{tab:os_loss} show that our proposed local orientation consistency loss force the model to aware and able to guide the model to correct these violations in resulting solution. Compare to the global regularization loss in VM in Table \ref{tab:result}, the proposed local orientation consistency maintains the resulting solutions to be diffeomorphic without exceedingly sacrificing the registration accuracy. 

	\begin{table}
		\centering	
		\renewcommand{\arraystretch}{0.95}
		\newcolumntype{C}[1]{>{\centering\arraybackslash}p{#1}}
		\begin{tabular}{C{1.4cm} C{1cm} C{1cm} C{1.4cm} C{1.3cm}}
			\toprule
			Time (s) & SyN & VM & DIF-VM & SYMNet\\
			\midrule
			Avg. &  1039 & 0.695 & 0.517 & 0.414 \\
			Std. & 59 & 0.381 & 0.121 & 0.012 \\
			\bottomrule
		\end{tabular}
		\vspace{8pt}
		\caption{Average and standard deviation of the running time in second for each deformable registration to register a pair of image volumes (lower is better).}\label{tab:time}
		\vspace{-8pt}
	\end{table}
	
	\subsubsection{Runtime Analysis}
	We report the average running time for non-linear deformable registration of each subject to an atlas using an Intel i7-7700 CPU and an NVIDIA GTX1080Ti GPU, where the running time for affine normalization is not included. Table \ref{tab:time} shows the average running time of the proposed methods and those baseline methods. It is worth noting that the implementation of SyN in ANTs utilizes CPU only, while the learning-based methods (\ie, DIF-VM, VM, SYMNet) utilize both CPU and GPU during deformable registration. We observe that learning-based methods are significantly faster than the traditional method SyN, which able to register a subject MR volume to an atlas with less than a second. The result shows that our method inherits the fast registration property form CNN-based registration methods. This implies that our method has a potential in real-time deformable registration in clinical applications. 
	
	
	\section{Conclusion}
	In this paper, we have presented a fast symmetric diffeomorphic approach to deformable image registration using CNN, which learns the symmetric deformation fields that align the pair of images to their mean shape within the space of diffeomorphic maps.  We have then proposed a novel local orientation-consistency loss that leverages the Jacobian determinant to further guarantee the desirable diffeomorphic properties of the resulting solutions. We have evaluated our model by using a large-scale brain MR dataset and compared our method to the classic registration approach and state-of-the-art unsupervised learning-based methods. The results obtained from the comprehensive experiments demonstrate that our method can outperform both the traditional method and learning-based methods in terms of registration accuracy and the quality of the deformation fields. 

	{\small
		\bibliographystyle{ieee_fullname}
		\bibliography{egbib}
	}

\end{document}